\documentclass[twoside,11pt]{article}

\usepackage{dmlr2e}

\usepackage{lastpage}
\dmlrheading{23}{2023}{1-\pageref{LastPage}}{9/23; Revised 10/23}{11/23}{21-0000}{Anonymous} 
\def\openreview{\url{https://openreview.net/forum?id=2OObXL3AaZ}} 

\ShortHeadings{Learning representations of learning representations}{Gonz\'alez-M\'arquez \& Kobak}
\firstpageno{1}

\usepackage{booktabs}
\usepackage{tabularx}
\usepackage{multirow}
\usepackage{enumitem}
\usepackage{wrapfig}

\begin{document}

\title{Learning representations of learning representations}

\author{\name Rita Gonz\'alez-M\'arquez \email rita.gonzalez-marquez@uni-tuebingen.de \\
       \addr Hertie Institute for AI in Brain Health,\\
       University of T\"ubingen, Germany
       \AND
       \name Dmitry Kobak \email dmitry.kobak@uni-tuebingden.de \\
       \addr Hertie Institute for AI in Brain Health,\\
       University of T\"ubingen, Germany; \\
       IWR, Heidelberg University, Germany}


\maketitle

\begin{abstract}
The ICLR conference is unique among the top machine learning conferences in that all submitted papers are openly available. Here we present the \textit{ICLR dataset} consisting of abstracts of all 24 thousand ICLR submissions from 2017--2024 with meta-data, decision scores, and custom keyword-based labels. We find that on this dataset, bag-of-words representation outperforms most dedicated sentence transformer models in terms of $k$NN classification accuracy, and the top performing language models barely outperform TF-IDF. We see this as a challenge for the NLP community. Furthermore, we use the ICLR dataset to study how the field of machine learning has changed over the last seven years, finding some improvement in gender balance. Using a 2D embedding of the abstracts' texts, we describe a shift in research topics from 2017 to 2024 and identify \textit{hedgehogs} and \textit{foxes} among the authors with the highest number of ICLR submissions.
\end{abstract}


\section{Introduction}
\label{sec:intro}

The International Conference on Learning Representations (ICLR) is one of the most prestigious machine learning venues: in Google Scholar Metrics it currently shares with NeurIPS the second place after CVPR. Since 2017, ICLR submissions are handled through OpenReview in a fully open way: all submitted papers are publicly visible and are eventually de-anonymized. This is not the case for most other top conferences in the field which do not make rejected papers openly visible. As the field of machine learning advances very fast, one can use ICLR submissions to study how it has changed over recent years.

\begin{figure}[b]
    \centering
    \includegraphics{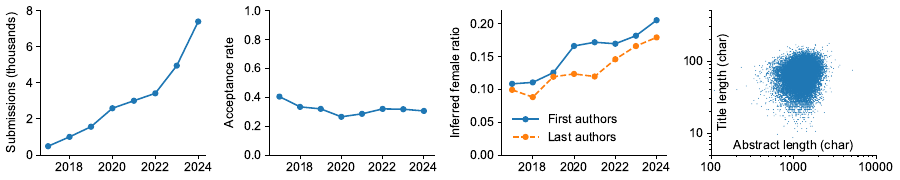}
    \caption{Summary statistics of the ICLR dataset (ICLR24v2).}
    \label{fig:summary-stats}
\end{figure}

Here we present the \textit{ICLR dataset} consisting of abstracts of all ICLR submissions from 2017--2024 with meta-data and keyword-based labels (Figure~\ref{fig:summary-stats}). Our work has two goals. First, to do a metascience study of machine learning as a field, similar to how \citet{gonzalez2023landscape} did it for biomedicine. Second, to frame an NLP challenge: without using our labels, train a language model that would substantially surpass a na\"ive TF-IDF representation in terms of $k$NN accuracy. We found that most dedicated sentence models fared \textit{worse} than TF-IDF, and none outperformed it by a large margin.

\section{Dataset}
\label{sec:data}

To assemble the dataset, we queried OpenReview and downloaded titles, abstracts, author lists, keywords, reviewers' scores, and conference decisions for all 24\,445 papers submitted to ICLR in 2017--2024 with intact abstracts (Figure~\ref{fig:summary-stats}). 26 submissions with placeholder abstracts (below 100 characters) were excluded. While 2024 submissions were still anonymous, we assembled the data into the ICLR24v1 dataset; after the de-anonymisation, we produced the final ICLR24v2 version. We will use the same naming convention for future updates. The data are openly available at \url{https://github.com/berenslab/iclr-dataset}, together with our analysis code.

We used the \texttt{gender} package \citep{genderpackage} to infer genders of the first and the last author based on their first names. We could infer genders for 41.8\% of the first authors and 49.9\% of the last authors; note that the inference model fails at inferring gender for many non-Western names. We observed a steady increase in the inferred female ratio (based only on papers with inferred genders) that almost doubled since 2017: from 11\% in 2017 to 21\% in 2024 for the first authors, and from 10\% to 18\% for the last authors.

To label the dataset, we relied on the author-provided keywords and used them to assign papers to 45 non-overlapping classes (Table~\ref{tab:classes}). We took the 200 most frequent keywords, combined some of them together into one class (e.g. \textit{attention} and \textit{transformer}), disregarded very broad keywords (e.g. \textit{deep learning}), and assigned papers to rarer keywords first. Using this procedure, we ended up labeling 53.4\% of all papers.

Reviewed papers had on average 3.7 reviews, with 93\% having either 3 or 4 reviews. Across all 244\,226 possible pairs of reviews of the same paper, the correlation coefficient between scores was 0.40. This was substantially higher than what had been reported for computational neuroscience conferences --- 0.16 for CCN \citep{tweetCCN} and 0.25 for Cosyne \citep{tweetCosyne}, --- but note that the ICLR scores are not entirely independent as the reviewers are allowed to update them after discussion.

\section{Embedding challenge}
\label{sec:challenge}

\begin{table}[t!]
\vspace{-.5em}
\caption{$k$NN accuracies in high-dimensional spaces, and in the 2D space after $t$-SNE. All values should be interpreted with an uncertainty of ${\sim}1\%$ (see text).}
\label{tab:acc}
\centering
\vspace{0.5em}
\begin{tabular}{lcc}
\toprule
\textbf{Model} & \textbf{High-dim.}  & \textbf{2D}\\ \midrule
TF-IDF & 59.2\% & 52.0\%\\
SVD & 58.9\% & 55.9\%\\
SVD, $L^2$ norm. & \textbf{60.7\%} & \textbf{56.7}\%\\
\midrule
SimCSE & 45.1\% & 36.3\%\\ 
DeCLUTR-sci & 52.7\%  & 47.1\%\\ 
SciNCL & 58.8\%  & 54.9\%\\ 
SPECTER2 & 58.8\%  & 54.1\%\\ 
ST5 & 57.0\%  & 52.6\%\\
SBERT & \textbf{61.6\%} & \textbf{56.8\%}\\ 
\midrule
Cohere v3 & 61.1\%  & 56.4\%\\ 
OpenAI v3 & \textbf{62.3\%} & \textbf{57.1}\%\\ 
\bottomrule
\end{tabular}
\end{table}

To obtain an embedding of each abstract, we used classic bag-of-words representations \citep{schmidt2018stable} as well as modern sentence transformers, pre-trained on large amounts of text data. We evaluated all of them using $k$NN classification accuracy ($k=10$ and 10-fold cross-validation). As our main application is 2D visualisation (Section~\ref{seq:trends}) which is based on the $k$NN graph, we consider $k$NN accuracy one of the most relevant metrics quantifying representation quality.

We used TF-IDF representation with log-scaling as implemented in scikit-learn \citep{sklearn}, which showed the best results in our prior benchmark \citep{gonzalez2022two}. Its $k$NN accuracy was 59.2\% (Table~\ref{tab:acc}). It decreased to 58.9\% after SVD to 100 dimensions, but increased to 60.7\% after $L^2$ normalisation in the SVD space (or, equivalently, using the cosine metric for $k$NN search).

As sentence transformers, we used three models which were specifically trained to produce representations of scientific abstracts: DeCLUTR-sci \citep{giorgi2021declutr}, SciNCL \citep{ostendorff2022scincl}, and SPECTER2 \citep{cohan2020specter}. We also used SimCSE \citep{gao2021simcse}, ST5 \citep{Ni2021SentenceT5}, and the latest version of SBERT \citep{reimers2019sbert}. The SBERT model (\texttt{all-mpnet-base-v2}) was trained on over one billion documents from different domains and holds state-of-the-art results in recent benchmarks among all models of its size \citep{muennighoff2023mteb}. These six models all have \texttt{bert-base} architecture with 110~M parameters and 768-dimensional embeddings. To get the representation of each abstract, we used the representation that each model had been fine-tuned for: either average pooling over all tokens (SBERT, DeCLUTER-sci, ST5) or the classification token \texttt{[CLS]} (SciNCL, SPECTER2, SimCSE). All models were downloaded from Hugging Face. We also benchmarked two commercial models: one by Cohere (\texttt{embed-english-v3.0} in \texttt{clustering} mode; 1024-dimensional embeddings), and one by OpenAI (\texttt{text-embedding-3-large}; 3072-dimensional embeddings). For all models we report $k$NN accuracy using the Euclidean metric; the cosine metric gave very similar results.

We found that the three models specifically trained to represent scientific abstracts all had lower $k$NN accuracy than TF-IDF. Only SBERT and both commercial models could outperform TF-IDF, and only marginally, by less than 2 percentage points (Table~\ref{tab:acc}). SBERT (61.6\%) was only surpassed by the OpenAI embedding model (62.3\%) with the performance gap below 1 percentage point. Note that all reported values should be interpreted with an error of around $\pm$1\%, corresponding to the binomial standard deviation $100 \sqrt{p(1-p)/n}$ for test set size $n\approx 2500$ and accuracy $p \approx 0.6$.

These results were surprising for us, because sentence transformers are complex models pre-trained with masked language modeling \citep{devlin2018bert} and fine-tuned with contrastive loss functions on large corpora. Yet their representations were not (much) better than bag-of-words representations that capture nothing beyond word counts. Modern benchmarks evaluate embedding models using various metrics and do find that sentence transformers outperform bag-of-words models \citep{muennighoff2023mteb}. However, the  $k$NN graph quality is the only metric relevant for our application (see below), and here the modern models were not much better than TF-IDF, at least on our dataset. 

We hope that this well-defined and practically relevant benchmark will act as a challenge for the NLP community.

\begin{figure}[t]
    \centering
    \includegraphics{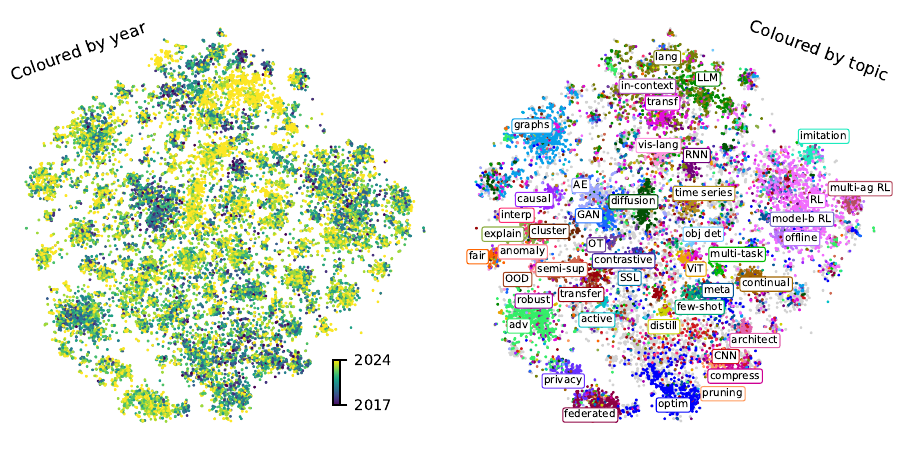}
    \caption{$t$-SNE embedding of the SBERT representation of ICLR abstracts (2017--2024). Left: coloured by year; right: coloured by topic.}
    \label{fig:embedding}
\end{figure}

\section{Trends in machine learning}
\label{seq:trends}

For data exploration, we used the SBERT representation and applied $t$-SNE \citep{vandermaaten2008tsne} to embed the 768-dimensional representation in 2D. We chose $t$-SNE rather than UMAP \citep{mcinnes2018umap} because $t$-SNE performs the best in terms of $k$NN classification and $k$NN recall ~\citep{gonzalez2022two,gonzalez2023landscape}. We used openTSNE \citep{polivcar2019opentsne} with default parameters. In 2D, $k$NN classification was 56.8\% (Table~\ref{tab:acc}): very close to what we got using TF-IDF (56.7\%) and using the OpenAI model (57.1\%).

The resulting embedding showed rich structure with many visible clusters roughly corresponding to our classes (Figure~\ref{fig:embedding}). Related classes were located close in the embedding, showing meaningful global organisation.

Overlaying the conference year and the class labels over the embedding (Figure~\ref{fig:embedding}) highlighted many trends in 2017--2024 machine learning. We saw that generative adversarial networks (GAN) and variational autoencoders (AE) got out of fashion while diffusion models became fashionable. Natural language processing research got dominated by large language models (LLM). Within reinforcement learning (RL), offline RL seemed to be the most recently fashionable topic. Recurrent neural networks (RNN) and adversarial examples are another two topics that lost their popularity.

We also used the 2D embedding to explore the distribution of acceptance decisions and average scores across machine learning subfields, but found no systematic differences between them (Figure~\ref{fig:acceptance}). This suggests that ICLR's decisions were not biased towards certain topics. Similarly, we did not see any systematic differences in gender ratio between machine learning subfields (Figure~\ref{fig:genders}), in stark contrast with biomedical research \citep{gonzalez2023landscape} and academia as a whole \citep{lariviere2013bibliometrics, shen2018persistent, bendels2018gender}.

\begin{figure}[t]
    \centering
    \includegraphics{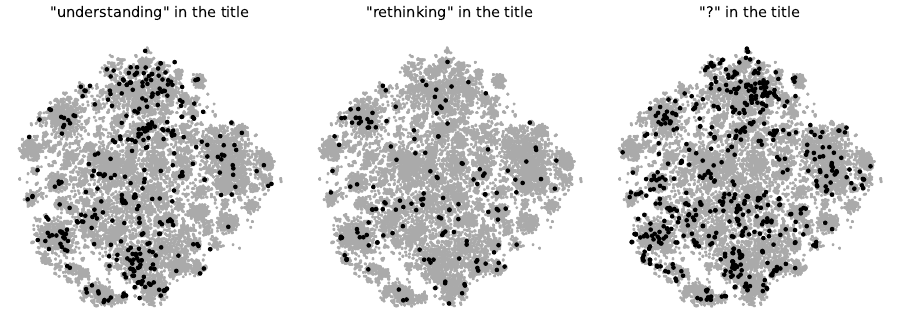}
    \caption{ICLR papers containing the words \textit{understanding} (366), \textit{rethinking} (155), and a question mark (550) in the title.}
    \label{fig:understanding}
\end{figure}

Which subfields of machine learning are the most controversial? We investigated this question by looking at the distribution of papers containing the words \textit{understanding}, \textit{rethinking}, or the question mark in their titles (Figure~\ref{fig:understanding}). These distributions were not uniform and had local modes around language models, vision-language models, adversarial examples, and also around optimisation/distillation.

\begin{figure}[t]
    \centering
    \includegraphics{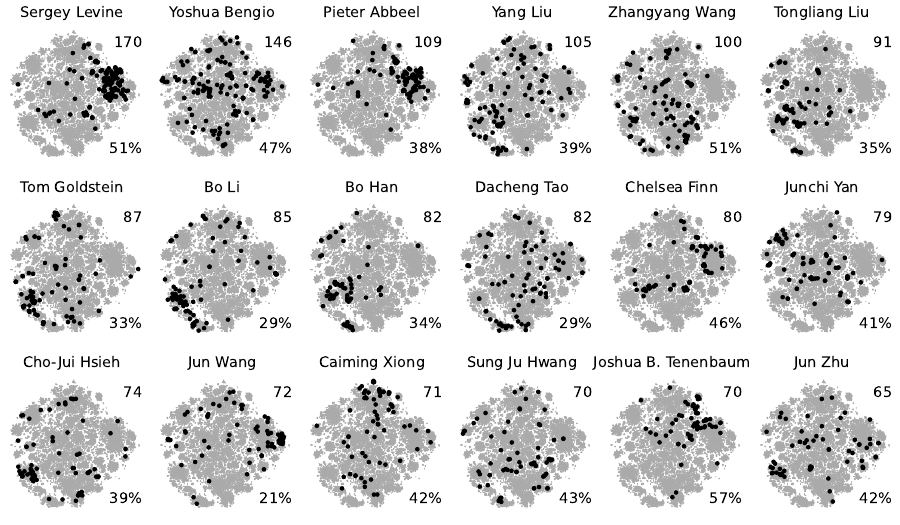}
    \caption{Top 18 authors by the total number of ICLR submissions over 2017--2024. Each panel shows the total number of submissions and the acceptance rate.}
    \label{fig:authors}
\end{figure}

Finally, we looked at the authors with the highest number of ICLR submissions (Figure~\ref{fig:authors}) and saw clear distinction between focused researchers working mostly on one topic and broad researchers working in many machine learning fields: \textit{hedgehogs} and \textit{foxes}, according to the famous classification by Isaiah Berlin (\citeyear{berlin1953hedgehog}). Among the top three most prolific authors, Sergey Levine (170 submissions) and Pieter Abbeel (109) were `hedghehogs' working mostly on reinforcement learning, while Yoshua Bengio (146) was a `fox'. The acceptance rate among the most prolific authors in both categories was often higher than the average acceptance rate (31\%).

\section{Conclusion}

Many text datasets are available for training and benchmarking language models. The benefits of the \textit{ICLR dataset} suggested here are (i) its compact size; (ii) it not being part of the training set of existing sentence transformer models; (iii) it covering topics very familiar to machine learning researchers, allowing qualitative assessment of embedding quality.

We demonstrated that the ICLR dataset can be used to study metascientific questions and to draw conclusions about the state of machine learning field as a whole. We also argue that substantially outperforming TF-IDF representation remains an open NLP challenge (Figures~\ref{fig:tfidf}, \ref{fig:openai}).

\vfill

\section*{Acknowledgements}

This research was funded by the Deutsche Forschungsgemeinschaft (KO6282/2-1 and Excellence clusters 2064 ``Machine Learning: New Perspectives for Science'', EXC 390727645, and 2181 ``STRUCTURES'', EXC 390900948), by the German Ministry of Education and Research (T\"ubingen AI Center), and by the Gemeinn\"{u}tzige Hertie-Stiftung. The authors thank the International Max Planck Research School for Intelligent Systems (IMPRS-IS) for supporting Rita Gonz\'alez M\'arquez.

\clearpage

\bibliography{main}

\clearpage
\appendix

\renewcommand{\thefigure}{S\arabic{figure}}
\setcounter{figure}{0}  
\renewcommand{\thetable}{S\arabic{table}}
\setcounter{table}{0} 
\renewcommand{\theHtable}{Supplement.\thetable}
\renewcommand{\theHfigure}{Supplement.\thefigure}

\section{Supplementary Figures}

\begin{figure}[h]
    \centering
    \includegraphics{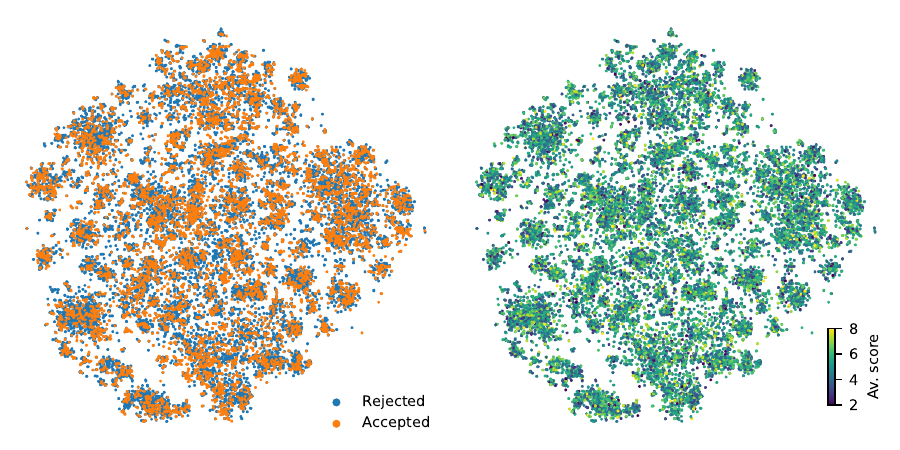}
    \caption{Acceptance decisions and average scores. Left: accepted papers are shown on top. Right: papers are shown in randomized order.}
    \label{fig:acceptance}
\end{figure}

\begin{figure}[h]
    \centering
    \includegraphics{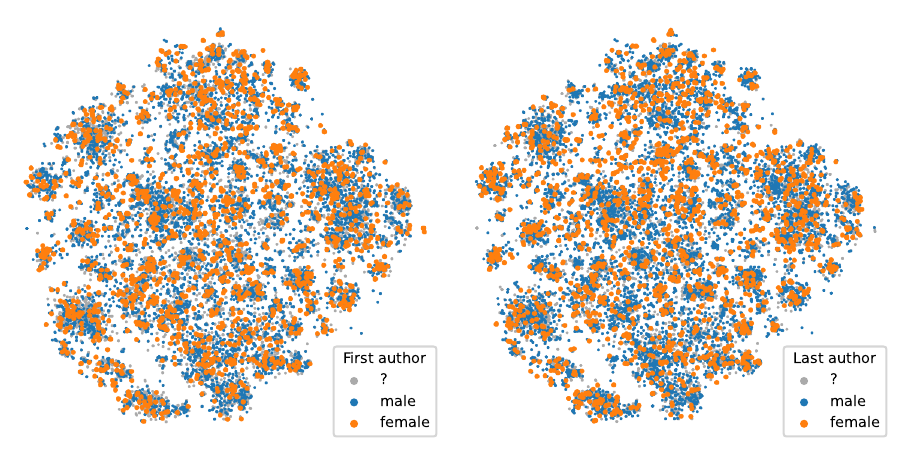}    
    \caption{Inferred genders of the first and the last authors. Papers are plotted in the following order: unknown gender, male, female. Female markers are larger.}
    \label{fig:genders}
\end{figure}

\begin{figure}[h]
    \centering
    \includegraphics{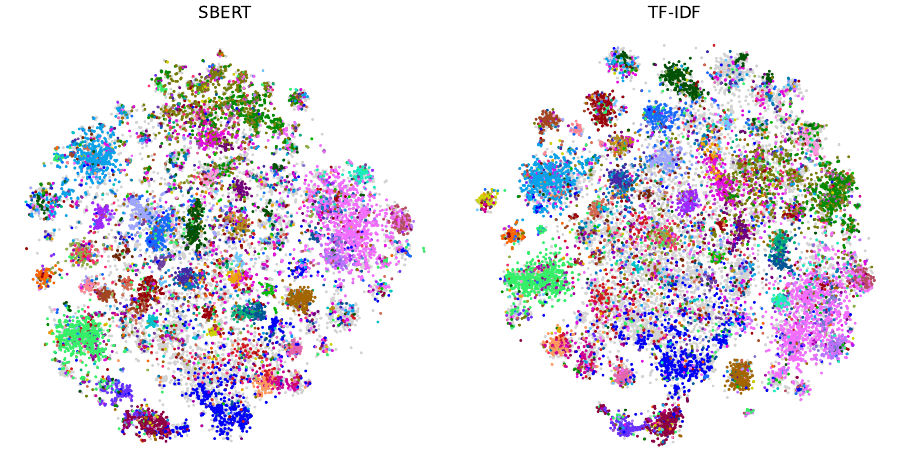}
    \caption{$t$-SNE embeddings of the SBERT representation (left) and of the row-normalized SVD (100 components) of the TF-IDF representation (right). The embedding on the right was rotated by 90$^\circ$ and flipped to align it to the SBERT embedding. Colours as in Figure~\ref{fig:embedding}. Unlabeled papers are shown in gray in the background.}
    \label{fig:tfidf}
\end{figure}

\begin{figure}[h]
    \centering
    \includegraphics{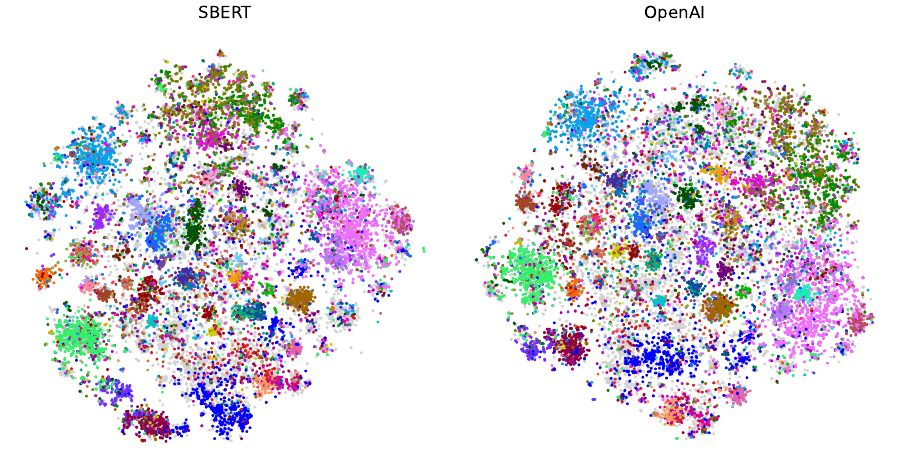}
    \caption{$t$-SNE embedding of the SBERT representation (left) and of the OpenAI's model representation (right). The embedding on the right was rotated by 90$^\circ$ and flipped. Colours as in Figures~\ref{fig:embedding} and ~\ref{fig:tfidf}.}
    \label{fig:openai}
\end{figure}

\clearpage
\section{Supplementary Tables}

\begin{table}[h]
\caption{The list of all 45 classes. From the list of all existing keywords, we selected the 200 most frequent ones and manually grouped some of them together into classes. We left out too general keywords (e.g. \textit{deep learning}) as well as all resulting classes with fewer than $50$ papers. Table continues on the next page.}
\label{tab:classes}
\centering
\tiny
\begin{tabularx}{.80\textwidth}{lcXc}
\toprule
\textbf{Class}      & \textbf{Samples}             & \textbf{Keyword}       & \textbf{Frequency}   \\
\midrule
\multirow{2}{*}{RL} & \multirow{2}{*}{1266} & reinforcement learning & 1608    \\ 
                      &   &  deep reinforcement learning & 298 \\ 
\midrule
\multirow{9}{*}{Adversarial} & \multirow{9}{*}{870} & adversarial training & 217    \\ 
                      &   &  adversarial attacks & 106 \\ 
                      &   &  adversarial defense & 50 \\ 
                      &   &  adversarial examples & 196 \\ 
                      &   &  adversarial learning & 93 \\ 
                      &   &  adversarial machine learning & 54 \\ 
                      &   &  adversarial robustness & 241 \\ 
                      &   &  adversarial & 60 \\ 
                      &   &  adversarial attack & 121 \\ 
\midrule
\multirow{8}{*}{Language models} & \multirow{8}{*}{802} & question answering & 59    \\ 
                      &   &  reasoning & 85 \\ 
                      &   &  language modeling & 85 \\ 
                      &   &  machine translation & 91 \\ 
                      &   &  language models & 151 \\ 
                      &   &  nlp & 166 \\ 
                      &   &  natural language processing & 433 \\ 
                      &   &  language model & 105 \\ 
\midrule
\multirow{9}{*}{Optimization} & \multirow{9}{*}{790} & optimization & 410    \\ 
                      &   &  gradient descent & 86 \\ 
                      &   &  combinatorial optimization & 69 \\ 
                      &   &  bayesian optimization & 64 \\ 
                      &   &  stochastic gradient descent & 77 \\ 
                      &   &  stochastic optimization & 56 \\ 
                      &   &  convex optimization & 57 \\ 
                      &   &  sgd & 86 \\ 
                      &   &  non-convex optimization & 66 \\ 
\midrule
\multirow{5}{*}{Graphs} & \multirow{5}{*}{730} & gnn & 64    \\ 
                      &   &  graph & 48 \\ 
                      &   &  graph representation learning & 85 \\ 
                      &   &  graph neural networks & 563 \\ 
                      &   &  graph neural network & 230 \\ 
\midrule
\multirow{5}{*}{Transformers} & \multirow{5}{*}{557} & transformer & 340    \\ 
                      &   &  self-attention & 73 \\ 
                      &   &  attention & 183 \\ 
                      &   &  attention mechanism & 53 \\ 
                      &   &  transformers & 261 \\ 
\midrule
\multirow{4}{*}{LLMs} & \multirow{4}{*}{538} & llm & 80    \\ 
                      &   &  prompting & 48 \\ 
                      &   &  large language model & 210 \\ 
                      &   &  large language models & 447 \\ 
\midrule
\multirow{3}{*}{Diffusion models} & \multirow{3}{*}{443} & diffusion models & 280    \\ 
                      &   &  diffusion model & 167 \\ 
                      &   &  diffusion & 69 \\ 
\midrule
\multirow{3}{*}{Transfer learning} & \multirow{3}{*}{419} & transfer learning & 388    \\ 
                      &   &  domain generalization & 124 \\ 
                      &   &  domain adaptation & 176 \\ 
\midrule
\multirow{4}{*}{GANs} & \multirow{4}{*}{380} & generative adversarial networks & 190    \\ 
                      &   &  gan & 168 \\ 
                      &   &  generative adversarial network & 70 \\ 
                      &   &  gans & 91 \\ 
\midrule
\multirow{5}{*}{Autoencoders} & \multirow{5}{*}{330} & variational autoencoders & 83    \\ 
                      &   &  autoencoders & 52 \\ 
                      &   &  autoencoder & 63 \\ 
                      &   &  variational autoencoder & 93 \\ 
                      &   &  vae & 71 \\ 
\midrule
\multirow{2}{*}{Continual learning} & \multirow{2}{*}{313} & lifelong learning & 82    \\ 
                      &   &  continual learning & 339 \\  
\bottomrule
\end{tabularx}
\end{table}

\begin{table}[h]    \centering
    \tiny
    \begin{tabularx}{.80\textwidth}{lcXc}
\toprule
\textbf{Class}      & \textbf{Samples}             & \textbf{Keyword}       & \textbf{Frequency}   \\
\midrule
\multirow{1}{*}{Federated learning} & \multirow{1}{*}{298} & federated learning & 485    \\ 
\midrule
\multirow{4}{*}{Out-of-distribution} & \multirow{4}{*}{275} & out-of-distribution generalization & 59    \\ 
                      &   &  distribution shift & 96 \\ 
                      &   &  out-of-distribution detection & 92 \\
                      &   &  out-of-distribution & 53 \\ 
\midrule
\multirow{2}{*}{Meta learning} & \multirow{2}{*}{275} & meta learning & 121    \\ 
                      &   &  meta-learning & 301 \\ 
\midrule
\multirow{1}{*}{Self-supervised learning} & \multirow{1}{*}{259} & self-supervised learning & 473    \\ 
\midrule
\multirow{4}{*}{RNNs} & \multirow{4}{*}{250} & lstm & 66    \\ 
                      &   &  recurrent neural networks & 114 \\ 
                      &   &  recurrent neural network & 48 \\ 
                      &   &  rnn & 65 \\ 
\midrule
\multirow{3}{*}{CNNs} & \multirow{3}{*}{247} & convolutional neural network & 76    \\ 
                      &   &  convolutional neural networks & 130 \\ 
                      &   &  cnn & 88 \\ 
\midrule
\multirow{1}{*}{Contrastive learning} & \multirow{1}{*}{244} & contrastive learning & 344    \\ 
\midrule
\multirow{2}{*}{Privacy} & \multirow{2}{*}{215} & differential privacy & 154    \\ 
                      &   &  privacy & 99 \\ 
\midrule
\multirow{2}{*}{Compression} & \multirow{2}{*}{214} & model compression & 135    \\ 
                      &   &  compression & 121 \\ 
\midrule
\multirow{3}{*}{Causality} & \multirow{3}{*}{202} & causal inference & 104    \\ 
                      &   &  causality & 80 \\ 
                      &   &  causal discovery & 53 \\ 
\midrule
\multirow{2}{*}{Explainability} & \multirow{2}{*}{194} & explainable ai & 92    \\ 
                      &   &  explainability & 131 \\ 
\midrule
\multirow{2}{*}{Offline RL} & \multirow{2}{*}{184} & offline rl & 55    \\ 
                      &   &  offline reinforcement learning & 150 \\ 
\midrule
\multirow{1}{*}{Interpretability} & \multirow{1}{*}{177} & interpretability & 356    \\ 
\midrule
\multirow{1}{*}{Semi-supervised learning} & \multirow{1}{*}{176} & semi-supervised learning & 253    \\ 
\midrule
\multirow{1}{*}{Robustness} & \multirow{1}{*}{175} & robustness & 411    \\ 
\midrule
\multirow{1}{*}{Few-shot learning} & \multirow{1}{*}{157} & few-shot learning & 218    \\ 
\midrule
\multirow{1}{*}{Multi-agent RL} & \multirow{1}{*}{151} & multi-agent reinforcement learning & 162    \\ 
\midrule
\multirow{1}{*}{Knowledge distillation} & \multirow{1}{*}{150} & knowledge distillation & 211    \\ 
\midrule
\multirow{1}{*}{Imitation learning} & \multirow{1}{*}{144} & imitation learning & 171    \\ 
\midrule
\multirow{2}{*}{Time series} & \multirow{2}{*}{140} & time series & 129    \\ 
                      &   &  time series forecasting & 54 \\ 
\midrule
\multirow{1}{*}{Neural architecture search} & \multirow{1}{*}{138} & neural architecture search & 180    \\ 
\midrule
\multirow{2}{*}{Pruning} & \multirow{2}{*}{133} & network pruning & 48    \\ 
                      &   &  pruning & 140 \\ 
\midrule
\multirow{1}{*}{Fairness} & \multirow{1}{*}{133} & fairness & 182    \\ 
\midrule
\multirow{1}{*}{Optimal transport} & \multirow{1}{*}{132} & optimal transport & 165    \\ 
\midrule
\multirow{2}{*}{ViTs} & \multirow{2}{*}{130} & vision transformers & 51    \\ 
                      &   &  vision transformer & 98 \\ 
\midrule
\multirow{1}{*}{Multi-task learning} & \multirow{1}{*}{121} & multi-task learning & 141    \\ 
\midrule
\multirow{1}{*}{Active learning} & \multirow{1}{*}{111} & active learning & 131    \\ 
\midrule
\multirow{2}{*}{Vision-language models} & \multirow{2}{*}{108} & vision-language models & 48    \\ 
                      &   &  clip & 70 \\ 
\midrule
\multirow{1}{*}{Object detection} & \multirow{1}{*}{106} & object detection & 125    \\ 
\midrule
\multirow{1}{*}{Model-based RL} & \multirow{1}{*}{105} & model-based reinforcement learning & 111    \\ 
\midrule
\multirow{1}{*}{Clustering} & \multirow{1}{*}{97} & clustering & 116    \\ 
\midrule
\multirow{1}{*}{Anomaly detection} & \multirow{1}{*}{87} & anomaly detection & 109    \\ 
\midrule
\multirow{1}{*}{In-context learning} & \multirow{1}{*}{87} & in-context learning & 105    \\  
\bottomrule
\end{tabularx}
\end{table}

\clearpage

\end{document}